\documentclass[runningheads]{llncs}

\usepackage{graphicx}
\usepackage{booktabs}
\usepackage{hyperref}
\usepackage{arxiv}

\newcommand\tp[2][-6]{{#2}^{\mkern#1mu\top}} 

\usepackage{xcolor}
\usepackage{array}

\usepackage{ulem,amssymb}
\begin{document}

\title{Forecasting Disease Progression with Parallel Hyperplanes in Longitudinal Retinal OCT}
\author{
  Arunava Chakravarty$^{1}$ \\
  Medical University of Vienna, Austria \\
  \texttt{arunava.chakravarty@meduniwien.ac.at}
  \And
  Taha Emre$^{1}$ \\
  Medical University of Vienna, Austria \\
  \texttt{taha.emre@meduniwien.ac.at}
  \And
  Dmitrii Lachinov$^{1}$ \\
  Medical University of Vienna, Austria \\
  \And
  Antoine Rivail$^{1}$ \\
  Medical University of Vienna, Austria \\
  \And
  Hendrik Scholl$^{2}$ \\
  University of Basel, Switzerland \\
  \And
  Lars Fritsche$^{3}$ \\
  University of Michigan, United States \\
  \And
  Sobha Sivaprasad$^{4}$ \\
  NIHR Moorfields Biomedical Research Centre, London, UK \\
  \And
  Daniel Rueckert$^{5}$ \\
  Technical University Munich, Germany \\
  \And
  Andrew Lotery$^{6}$ \\
  University of Southampton, United Kingdom \\
  \And
  Ursula Schmidt-Erfurth$^{1}$ \\
  Medical University of Vienna, Austria \\
  \And
  Hrvoje Bogunovi\'c$^{1}$ \\
  Medical University of Vienna, Austria \\
  \texttt{hrvoje.bogunovic@meduniwien.ac.at}
}
%\undertitle{The final version of this Preprint is accepted at MICCAI 2024,  \url{https://papers.miccai.org/miccai-2024/339-Paper3334.html}}
%\title{Forecasting Disease Progression with Parallel Hyperplanes in Longitudinal Retinal OCT}
%\author{Arunava Chakravarty$^{1}$ \and Taha Emre$^{1}$ \and Dmitrii Lachinov$^1$ \and Antoine Rivail$^1$ \and Hendrik Scholl$^{2}$ \and Lars Fritsche$^3$ \and Sobha Sivaprasad$^4$ \and Daniel Rueckert$^{5}$ \and Andrew Lotery$^6$ \and Ursula Schmidt-Erfurth$^1$ \and Hrvoje Bogunovi\'c$^1$}
%\authorrunning{A. Chakravarty et al.}
%\titlerunning{Forecasting Disease Progression with Parallel Hyperplanes}
%\institute{Dept. of Ophthalmology and Optometry, Medical University of Vienna, Austria \and
%Department of Ophthalmology, University of Basel, Switzerland \and University of Michigan, United States \and NIHR Moorfields Biomedical Research Centre, Moorfields Eye Hospital NHS Foundation Trust, London, United Kingdom \and
%Institute for AI and Informatics in Medicine, Klinikum rechts der Isar, Technical University Munich, Germany \and
%Clinical and Experimental Sciences, Faculty of Medicine, University of Southampton, United Kingdom \\
%\email{\{arunava.chakravarty,taha.emre,hrvoje.bogunovic\}@meduniwien.ac.at}}
\maketitle              % typeset the header of the contribution

\begin{abstract}
%Predicting future risk of disease progression from medical images is challenging due to varied progression speeds between individuals, and the biomarkers are often subtle and unknown clinically. 
Predicting future disease progression risk from medical images is challenging due to patient heterogeneity, and subtle or unknown imaging biomarkers. Moreover, deep learning (DL) methods for survival analysis are susceptible to %inter-scanner variations, leading to significant 
image domain shifts across scanners. We tackle these issues in the task of predicting late dry Age-related Macular Degeneration (dAMD) onset from retinal OCT scans. %\TEc{the previous sentence seems unnecessary, it only says we tackle}. 
We propose a novel DL method for survival prediction to jointly predict from the current scan a risk score, inversely related to time-to-conversion, and the probability of conversion within a time interval $t$. It uses a family of parallel hyperplanes generated by parameterizing the bias term as a function of $t$. In addition, %Using longitudinal scans during training, 
we develop unsupervised losses based on intra-subject image pairs to ensure that risk scores %from scans of the same eyes 
increase over time and that future conversion predictions are consistent with AMD stage prediction using actual scans of future visits. Such losses enable data-efficient fine-tuning of the trained model on new unlabeled datasets acquired with a different scanner.
Extensive evaluation on two large datasets acquired with different scanners %with significant domain shift, 
%highlights the benefit of our method for personalized management and clinical trial recruitment. 
resulted in a mean AUROCs of 0.82 for Dataset-1 and 0.83 for Dataset-2, across prediction intervals of 6,12 and 24 months.
%\TE{with limited fine-tuning}\TEd{(after fine-tuning with only $25\%$ of unlabeled training data)} 
\end{abstract}

\section{Introduction}
Predicting the risk of disease progression is essential for prioritizing high-risk patients for timely treatment and clinical trial recruitment. However, this task is challenging due to several factors. First, the lack of well-established clinical biomarkers makes it difficult to predict future disease progression. Second, %\TEd{data censoring can lead to unknown time-to-conversion labels due to missing follow-ups or lack of the conversion onset within the study period.}
missing follow-ups or lack of the conversion onset within the study period can lead to unknown time-to-conversion labels. %does not occur within \TEd{the study's limited duration}\TE{the study period}. 
Third, only a small proportion of monitored patients actually undergo conversion, resulting in severly imbalanced datasets.
%Third, \TE{severe} class imbalance \TEd{is an issue}, as only a small proportion of monitored patients actually undergo conversion, \TEd{resulting}\TE{results} in imbalanced datasets\TEc{class imbalance causes imbalanced dataset?}. 
Finally, discretizing time into bins for conversion prediction poses challenges such as having imprecise labels during training and the inability to predict conversions at arbitrary continuous times during inference.
Inter-scanner variations in intensity and noise profiles among different scanner manufacturers and image acquisition settings can result in significant domain shifts \cite{nyul2000new}.  Consequently, there is often a need to fine-tune existing model weights trained on images from one scanner to work on images from the other ones. However, the availability of a limited amount of images for fine-tuning and the absence of manual annotations often pose challenges. This highlights the need for exploring innovative methods to fine-tune existing models using limited unlabeled data.

In this work, we address these issues in the context of Age-Related Macular Degeneration (AMD), a leading cause of blindness among the elderly population~\cite{wong2014global}. While asymptomatic in its early and intermediate stages (iAMD), characterized by drusen, it progresses to a late stage that can be either dry (dAMD) or wet (nAMD), resulting in irreversible vision loss. dAMD is more prevalent, marked by Geographic Atrophy (GA). %with a loss of retinal pigment epithelium (RPE) and the photoreceptor layer. 
With recent FDA approvals for drugs to treat dAMD \cite{Khanani2023,Heier2023}, regular monitoring of eyes in the iAMD stage using longitudinal Optical Coherence Tomography (OCT) imaging is crucial to initiate treatment at the earliest onset of dAMD and minimize vision loss.
%%%%%%%%%%%%%%%%%%%%%%%%%%%%%%%%%%%%%%%%%%%%%%%%%%%%%

Existing methods for forecasting iAMD to nAMD or dAMD conversion can be categorized into biomarker and image-based approaches. \textit{Biomarker-based} methods involve segmenting retinal tissues and combining handcrafted features with clinical and demographic data \cite{sleiman2017optical,schmidt2018prediction,banerjee2020prediction,de2014quantitative,lad2022machine}. For example, \cite{banerjee2020prediction} employs biomarkers from past visits in an LSTM network for risk assessment. \textit{Image-based} methods utilize deep learning (DL) models on raw OCT scans, bypassing manual segmentation. Unlabeled longitudinal OCT datasets have been utilized for feature learning via temporal self-supervised learning \cite{emre2022tinc,rivail2019modeling}. These methods often employ binary or multi-label classification for predicting conversion within specific timeframes, such as 2 years \cite{russakoff2019deep} or 6-12-18 months \cite{rivail2019modeling}, with limited handling of censoring. A hybrid approach using both biomarker and image features was explored in \cite{yim2020predicting}. Survival analysis addresses challenges like censoring \cite{rivail2023deep}, using traditional CoxPH models \cite{schmidt2018prediction}. DL extensions such as DeepSurv remain unexplored in AMD progression \cite{katzman2018deepsurv}. Parametric models like CoxPH are inflexible, which neural-ODE-based methods such as SODEN attempt to overcome \cite{tang2022soden}. Recently, N-ODEs have been applied to model GA growth from OCT images \cite{lachinov2023learning} and Diabetic retinopathy progression from fundus images \cite{zeghlache2023lmt}.
%%%%%%%%%%%%%%%%%%%%%%%%%%%%%%%%%%%%%%%%%%%%%%%%%%%%%%

\textbf{Our Contributions:} 
(i) We propose a novel method for forecasting disease progression in continuous time using a family of parallel hyperplanes $\mathcal{H}(t)$. Each $\mathcal{H}(t)$  divides the feature space into two half-spaces:  one with samples not converting within the next $t$ months, and the other with samples converting to dAMD within $t$.
(ii) Our method jointly predicts both a conversion risk score which is inversely related to the conversion time as well as the Cumulative distribution function (CDF) of the conversion time. This risk score $r$ aids in stratifying patients into different risk groups.
(iii) We explore a way to fine-tune our model for adapting it across different imaging scanners with limited, unlabelled training images. Leveraging longitudinal pairs of scans for each eye, we employ unsupervised losses based on intra-subject consistency. These ensure that the predicted conversion probability at future time-points is consistent with the AMD stage prediction obtained from actual OCT scans of future visits. Additionally, we incorporate a ranking loss on predicted risk scores to ensure conversion risk consistently increases for future visits, as AMD is a degenerative disease that only progresses with time.
(iv) Extensive evaluation is performed on multi-center and multi-scanner datasets exhibiting a significant image domain shift.
%\TEd{Extensive evaluation is performed using datasets from different sites and scanners}\TE{Extensive evaluations are conducted on datasets from different sites and scanners}, representing\TEc{revealing?} a significant domain shift.
%%%%%%%%%%%%%%%%%%%%%%%%%%%%%%%%%%%%%%%%%%%%%%%%%%%%%%%%%
\section{Method}

\begin{figure}[!tb]
 \centering
  \includegraphics[width=.99\textwidth]{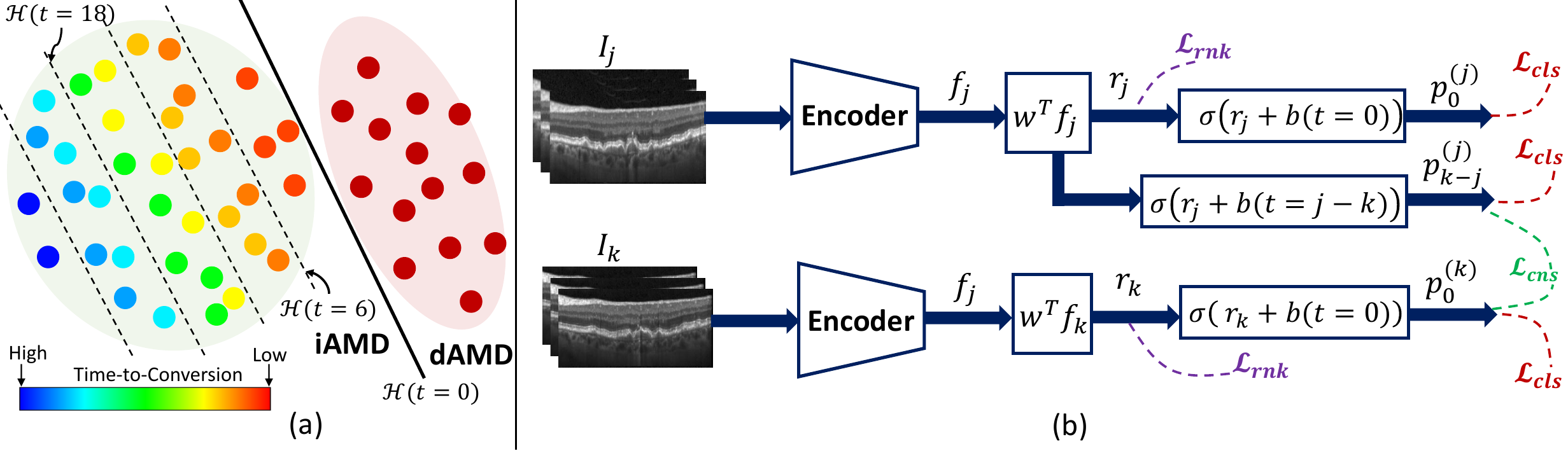}
\caption{(a) Illustration of the proposed method. (b) Proposed Training pipeline employs the same ConvNeXt-Tiny Encoder in both branches with shared weights. }
\label{fig:method} 
 \end{figure}
 
% Disease Forecasting: ranking and CDF prediction tasks
\textbf{Proposed Disease Progression Formulation:}
The proposed method combines two distinct approaches of forecasting disease progression. First, we model the conversion time as a random variable $T^*$ with corresponding Cumulative Distribution Function (CDF) function $p(t|\vec{I})=P(T^*\le t|\vec{I})$ and $\vec{I}\in\mathcal{I}$ is a space of all possible images. Second, for each $\vec{I}$ with conversion time $T$ we assign a risk score $r:\mathcal{I}\rightarrow\mathbb{R}$ such that $\forall \vec{I_1},\vec{I_2}\in\mathcal{I}, T_1\leq T_2 : r(\vec{I_1})\geq r(\vec{I_2})$. The first interpretation estimates a patient's $t$-year survival likelihood (CDF) while the second can stratify a population into low, medium, and high risk groups  by thresholding the risk score.
%Forecasting disease progression can be accomplished in two ways. First, given the current scan $\vec{I}$, model the Cumulative Distribution Function (CDF) $p_t=P(T^*\le t)$, where $T^*$ represents the conversion time. $p_t$ is the probability that $\vec{I}$ will convert to the advanced dAMD stage within the given time-interval t from the current visit. An alternative survival prediction task is to predict a scalar risk score $r$ from $\vec{I}$ that can rank the input scans in inverse order of their conversion times. These two tasks have clinical applications, such as predicting a patient's $t$-year survival likelihood (CDF) or stratifying a population into low, medium, and high risk groups (ranking) by thresholding the risk score (see Appendix xx). 
% iAMD vs nAMD classifier
Our proposed formulation illustrated in Fig. \ref{fig:method}(a) combines both of these approaches.

A CNN encoder maps each scan $\vec{I}$ to a point in the feature embedding space (represented as dots in Fig. \ref{fig:method}(a)). Then, these features are fed into a linear \textit{iAMD vs. dAMD} stage classifier with weights $\vec{w}$ and a scalar bias $\beta$. Notably, the $\vec{w}$ vector is \textit{normal} to the classifier’s decision hyperplane $H$ such that $\forall f \in H: \tp[-4]{\vec{w}}\vec{f}+ \beta=0$. For a given $\vec{f}$, its shortest signed distance perpendicular to $H$ is proportional(scaled by a factor of $||w||$) to $\tp[-4]{\vec{w}}.\vec{f}+ \beta$. The iAMD samples lie on the negative half-space ($\tp[-4]{\vec{w}}.\vec{f}+ \beta <0$) with negative signed distances from $H$.

%The coordinate of $f$\HBc{where is f defined and should it be bold here?} in projective space $Span\{w\}$, which we call a signed distance from $H$, is smaller than $\beta$ for iAMD cases and larger for dAMD.
%is proportional to the signed distance (scaled by $||w||$) of $\vec{f}$ from $H$ \TEc{what is $\vec{f}$?}. \TEc{Ideally?} The iAMD samples lie on the negative half-space (with negative signed distances from $H$). 

% Temporal ranking
\textbf{Risk score based on distance from $H$}: We introduce a temporal ordering among the iAMD samples with a ranking loss $\mathcal{L}_{rnk}$ (details discussed below), such that the distance of each sample from $H$ is inversely related to its conversion time $T^*$. This is illustrated in Fig.~\ref{fig:method}(a) by color grading each iAMD sample from red to blue in increasing order of $T^*$. Rank ordering serves multiple objectives. First, it acts as a regularizer for learning a semantically meaningful feature space with a better chance of generalization when trained on limited labeled data. Second, the signed distance from $H$ can be used as a risk score 
%\vspace{-5pt}
\begin{equation}
r=\tp[-4]{\vec{w}}\vec{f},
\label{eqn_risk}
\end{equation}
%\vspace{-1pt}
such that the higher the $r$, the closer it is to conversion.  $\beta$ being constant across all samples can be ignored for ranking. Risk score is further calibrated in a post-processing step to normalize to $[0,1]$. After training, $r$ is obtained for all scans in the validation set and a bicubic interpolation is learned to map the k-th percentile of $r$ values to $k/100$ in increments of 10 percentiles.

%A further calibration step is employed as a post-processing step to map $r$ to $[0,1]$ with bicubic interpolation based on the percentile values of the predicted risk scores on the validation set.
% After training, r is obtained for all scans in the validation set. We learn a bicubic interpolation which maps k-th percentile of r values to k/100 in increments of 10 percentiles 

\textbf{Modeling CDF with a family of hyperplanes parallel to $H$}: 
% How to model CDF?
We propose a novel continuous-time modeling of the CDF to leverage the temporal ranking in the feature space. Predicting $p_t$ involves learning a separate linear binary classifier for each time-interval $t$ to predict if $\vec{I}$ will convert within $t$. $t$ is normalized such that $0-3$ years is linearly mapped in the range $[0,1]$. We extend our formulation by considering a continuous family of separating hyperplanes  $\mathcal{H}(t)$ parallel to $H$, each predicting the conversion within $t$ (depicted with dashed lines in Fig. \ref{fig:method}(a)). All hyperplanes in $\mathcal{H}(t)$ share the normal vector $\vec{w}$ but employ a different bias $b(t)$, parameterized as a monotonic function of $t$. Thus, 
%\vspace{-5pt}
\begin{equation}
p_t(\vec{I})=\sigma \left( \tp[-4]{\vec{w}}\vec{f}+b(t) \right)=\sigma \left( r + b(t) \right),
\label{eqn_cdf}
\end{equation}
%\vspace{-1pt}
where $\sigma(.)$ is the sigmoid function. We reformulated bias $b(t)$ as an affine transformation $b(t)=\alpha\cdot t + \beta$, where $\alpha$ and $\beta$ are scalar learnable parameters of our model. The hyperplane $H$ for iAMD vs dAMD stage classification is a member of this family, $H=\mathcal{H}(t=0)$. Thus, $p_{0}=\sigma \left( \tp[-4]{\vec{w}}\vec{f}+b(0) \right)=\sigma \left( \tp[-4]{\vec{w}}\vec{f}+ \beta \right)$ is the probability that the current input scan $\vec{I}$ has already converted to the dAMD.

\textbf{Training Pipeline: }
%%%%%%%%%%%%%%
% Intra-subject consistency
%In addition to the temporal ranking by the conversion time, the availability of longitudinal data can be leveraged to incorporate additional regularization of the feature embedding. We consider a pair of $\vec{I}_j$, $\vec{I}_k$ of the same eye imaged at time-points $j$ and $k$ respectively, such that $\vec{I}_j$ precedes $\vec{I}_{k}$ with $j<k$.
%Since degenerative diseases such as AMD can only progress (and not regress) with time, the predicted risk scores their risk scores should follow $r_j<r_k$
We consider a Siamese architecture (Fig. \ref{fig:method}(b)) during training to leverage the availability of longitudinal images by considering image-pairs from different eyes in each training batch. Each random image-pair $(\vec{I}_j, \vec{I}_k)$ are two OCT scans of the same eye, acquired at different patient visits at time-points $j$ and $k$. $\vec{I}_j$ precedes $\vec{I}_k$ (i.e., $j<k$) with a time-interval of (j-k) between them. Both $\vec{I}_j, \vec{I}_k$ are fed to an Encoder (ConvNeXt-Tiny initialized with ImageNet pretrained weights \cite{liu2022convnet}), to obtain the features $\vec{f}_j, \vec{f}_k$ respectively. Their risk scores $r_j$ and $r_k$ are obtained using Eq. \ref{eqn_risk} and the probabilities $p_{0}^{(j)}$ and $p_{0}^{(k)}$ that $\vec{I}_j$ and $\vec{I}_k$ have already converted to dAMD are computed using Eq. \ref{eqn_cdf}. Additionally, we compute the probability of $\vec{I}_j$ to convert to dAMD within the next (k-j) time-interval as $p_{k-j}^{(j)}=\sigma \left(r_j+b(k-j) \right)$. 
Thus, while $p_{j}^{(0)}$ and $p_{k}^{(0)}$ essentially perform an iAMD vs. dAMD stage classification for the input scan,  $p_{k-j}^{(j)}$ forecasts the conversion probability for a future time-point $k$, directly from a previous visit $\vec{I}_j$ without accessing $\vec{I}_k$.

\textbf{Loss Functions: }
Following survival analysis, the Ground Truth (GT) labels for each scan $\vec{I}_j$ is denoted by the tuple $\left( T_j, E_j \right)$. If the binary event indicator $E_j=1$, then the eye to which $\vec{I}_j$ belongs, converts to dAMD after a time-interval $T_j$ from the current visit.  $E_j=0$ indicates that the eye did not convert within the monitoring period in which case  $T_j$ represents time duration from the current to the last visit in the study after which the eye is \textit{censored}.

\textit{Classification Loss $\mathcal{L}_{cls}$}: The GT for iAMD vs dAMD classification for an eye at a time-point $j$ is given by $y_j=1$ if it has already converted, i.e., $E_j=1$ and $T_j<=0$, otherwise $y_j=0$. The binary cross entropy loss $\mathcal{L}_{bce}(.)$ is used to define the classification loss as $\mathcal{L}_{cls}=\mathcal{L}_{bce}(y_j, p^{(j)}_{0}) + \mathcal{L}_{bce}(y_k, p^{(k)}_{0}) +  \mathcal{L}_{bce}(y_k, p^{(j)}_{k-j})$.

\textit{Intra-Subject Consistency Loss $\mathcal{L}_{cns}$}: For a given eye, the conversion probability at a time-point $k$ predicted from the scan acquired at time $k$($p_{0}^{(k)}$) should be consistent with the probability forecast for $k$, using a previous scan from time-point $j$ ($p_{k-j}^{(j)}$). This is ensured with the consistency loss $\mathcal{L}_{cns}=\mathcal{L}_{bce}( p_{0}^{(k)}, p_{k-j}^{(j)} )$.
%%%%%%%%%%%%%%%%%%%%%%%%%%%%%%%%%%%%%%%%%%%%%%%%%%%%%%%%%%

\textit{Temporal Ranking Loss $\mathcal{L}_{rnk}$}: We consider all possible image-pairs $(\vec{I}_m, \vec{I}_n)$ in a training batch (including pair of scans coming from different eyes). $\mathcal{L}_{rnk}$ solves a logistic regression task using the difference in risks $(r_m-r_n)$ as input to a linear classifier to  predict the probability $p_{m<n}$ of $\vec{I}_m$ converting before $\vec{I}_n$ as $\mathcal{L}_{rnk}=- \frac{1}{|S_{m<n}|+|S_{m>n}|}. \left[ \sum_{S_m<n} log \left(p_{m<n} \right) +  \sum_{S_{m>n}} log \left(1-p_{m<n} \right) \right]$, 
where $S_{m<n}$ represents a subset of all possible image-pairs in a training batch where $I_m$ converts before $I_n$ for which ideally, $p_{m<n} \approx 1$. Similarly, $S_{m>n}$ contains image-pairs where $\vec{I}_m$ converts after $\vec{I}_n$ and ideally, $p_{m<n} \approx 0$. The set $S_{m<n}$ comprises image-pairs for which $\lbrace (T_m<T_n)$ $\&$ $(E_m=1$ or $\vec{I_m}$, $\vec{I_n}$ belong to the same eye)$\rbrace$. 
AMD progression is irreversible and the retinal tissue damage only accumulates with time. Therefore, even for cases where $E=0$ and the actual conversion time is unknown, the risk score of a scan from a later visit $\vec{I}_m$( with a smaller time duration $T_m$ to the last visit) should always be higher than a former visit $\vec{I}_n$ ($T_n>T_m$) of the same eye. Similarly, $S_{m>n}$ is defined as pairs where 
$\lbrace (T_m>T_n)$ $\&$ $(E_n=1$ or $\vec{I_m}$, $\vec{I_n}$ belong to the same eye)$\rbrace$.

%%%%%%%%%%%%%%%%%%%%%%%%%%%%%%%%%%%%%%%%%%%%%%%%%%%%%%%%%%%%

Thus, the total loss is defined as $\mathcal{L}=\mathcal{L}_{cls}+\mathcal{L}_{cns}+ \mathcal{L}_{rnk}$ with equal weights given to each term. While a Siamese two-branch architecture is employed during training, only a single branch is employed during inference. The proposed method employs a single visit's scan $\vec{I}$ as input to predict $r$ (see Eq. \ref{eqn_cdf}) and the probability of conversion $p_t$ within a given future time-interval $t$ (see Eq. \ref{eqn_risk}).

\textbf{Unsupervised Fine-tuning on External Datasets}: 
%Considerable domain shift across OCT scanners necessiates a fine-tuning step to adapt the model to new datasets coming from a different scanners requring additional training data with manually labeled GT. 
We adapt our training losses to facilitate unsupervised fine-tuning with unlabeled data.  $\mathcal{L}_{cls}$ requires GT labels and, therefore, is not used. The unsupervised loss $\mathcal{L}_{cns}$  leverages the consistency in the predictions from the two branches of the Siamese architecture and is retained unmodified. $\mathcal{L}_{rnk}$ is adjusted in how $\vec{I_m, \vec{I}_n}$ pairs are constructed within each training batch. In the absence of conversion time labels, risk scores of scans across the batch samples cannot be compared. Only the intra-subject sample pairs $\vec{I_j, \vec{I}_k}$ are utilized, as they should still be be ranked as $r_k> r_j$ for time-points $k>j$ as AMD being degenerative cannot regress with time.

\textbf{Implementation Details: }
Our method was implemented in Python 3.8, PyTorch 2.0.0 (code available at \url{https://github.com/arunava555/Forecast_parallel_hyperplanes} ). %utilizing 6GB of GPU memory for training with a batch size of 16. 
The \textit{training} comprised 200 epochs (with 300 batch updates per epoch, batch size of 16), employing the AdamW optimizer \cite{loshchilov2018decoupled} with a cyclic learning rate \cite{smith2017cyclical} varying between $10^{-6}$ to $10^{-4}$. Each training batch was constructed with random image-pairs $(\vec{I}_{j}, \vec{I}_k)$ with a time-interval of 0-3 years between them, ensuring that all $\vec{I}_{j}$ were in the iAMD stage, while half of the $\vec{I}_k$ in each batch were in the dAMD stage (through oversampling). Three consecutive B-scans (slices) out of the 5 central B-scans in the OCT volumes were randomly extracted and input to Encoder in place of the three RGB color channels. Data augmentations included random translations, horizontal flip, random crop-resize, Gaussian noise, random in-painting and random intensity transformations. During \textit{inference}, for each scan, three sets of 3-channel input images were formed from the 5 central OCT slices, each containing the central slice in the left, middle or right channel. An average of their predictions were used for all experiments (including the benchmark methods for comparison).

% real-world retrospective cohort of OCT scans from the PINNACLE consortium [19] collected from the University Hospital Southampton and Moorfields Eye Hospital.

\section{Experiments and Results}
\label{sec:exp}
\textbf{Datasets:} We comprehensively evaluated our method on \textit{Dataset-1} collected at the Department of Ophthalmology, Medical University of Vienna, comprising 3,534 OCT scans from 235 eyes (40 converters and 195 censored) acquired with a Spectralis OCT scanner at a resolution of 49 B-scans (slices), each with a $512-1024 \times 49$ px. For converter eyes, labels for each scan were computed by measuring the time interval between its acquisition and the first conversion visit.  We considered an additional independent, external real-world dataset, \textit{Dataset-2}, collected from two different sites (University Hospital Southampton and Moorfields Eye Hospital) from the PINNACLE consortium \cite{sutton2023developing}. It comprises a randomly divided training set of 254 eyes (2428 scans) with 49 converters; a validation set of 127 eyes (1073 scans) with 26 converters; and a test set of 254 eyes (2305 scans) with 49 converters. All scans were acquired with Topcon scanners at a resolution of 128 B-scans with a $885 \times 512$ px. The scans from \textit{Dataset-1} and \textit{Dataset-2} exhibit large image domain shift due to different imaging scanners.

\newcolumntype{H}{>{\setbox0=\hbox\bgroup}c<{\egroup}@{}}
\setlength{\tabcolsep}{6pt}
\renewcommand{\arraystretch}{1.2}
\begin{table}[]
\centering
\label{Tab:dataset1}
\caption{Ablation and Comparison with state-of-the art on \textit{Dataset-1}. %(mean $\pm$ std. deviation). Best values in each column is highlighted.
}
\resizebox{1.0 \textwidth}{!}{
\begin{tabular}{@{}l|llHl|llHl|l@{}}
\toprule
  & \multicolumn{4}{c|}{AUROC}   & \multicolumn{4}{c|}{Balanced Accuracy} &  \\ 
                                 & \multicolumn{1}{c}{6} & \multicolumn{1}{c}{12} & \multicolumn{1}{H}{18} & \multicolumn{1}{c|}{24} & \multicolumn{1}{c}{6} & \multicolumn{1}{c}{12} & \multicolumn{1}{H}{18} & \multicolumn{1}{c|}{24} & \multicolumn{1}{c}{CcI} \\ \midrule

%\multicolumn{10}{c}{Ablation} \\
%\midrule
$\mathcal{L}_{cls}$  & $0.816\pm0.05$& $0.792\pm0.06$& $0.766\pm0.06$& $0.771\pm0.06$& $0.802\pm0.04$& $0.769\pm0.05$& $0.747\pm0.04$& $0.744\pm0.04$ & $0.740\pm0.06$   \\ 
$\mathcal{L}_{cls} + \mathcal{L}_{rnk}$ & $0.823\pm0.07$& $0.805\pm0.04$& $0.780\pm0.02$& $0.772\pm0.02$& $\mathbf{0.816\pm0.05}$& $0.772\pm0.02$& $0.751\pm0.02$& $0.742\pm0.02$ & $0.752\pm0.03$  \\ \midrule

Proposed  & $\mathbf{0.825\pm0.09}$& $\mathbf{0.828\pm0.07}$& $\mathbf{0.815\pm0.05}$& $\mathbf{0.809\pm0.06}$& $0.813\pm0.06$& $\mathbf{0.798\pm0.05}$& $\mathbf{0.780\pm0.05}$& $\mathbf{0.770\pm0.06}$ & $\mathbf{0.783\pm0.06}$ \ \\ 
\midrule

%\multicolumn{10}{c}{Scan-level performance.} \\ \midrule  
Cens. CE \cite{wulczyn2020deep} & $0.787\pm0.06$& $0.779\pm0.06$& $0.776\pm0.05$& $0.789\pm0.04$& $0.764\pm0.05$& $0.739\pm0.04$& $0.731\pm0.03$& $0.741\pm0.02$ & $0.767\pm0.04$   \\ 
 
Logis. Hazard~\cite{rivail2023deep}    & $0.787\pm0.06$& $0.787\pm0.04$& $0.779\pm0.04$& $0.797\pm0.03$& $0.780\pm0.06$& $0.766\pm0.03$& $0.745\pm0.04$& $0.755\pm0.04$ & $0.769\pm0.04$ \\ 
DeepSurv  \cite{katzman2018deepsurv} & $0.755\pm0.13$& $0.735\pm0.12$& $0.720\pm0.11$& $0.728\pm0.12$& $0.734\pm0.12$& $0.702\pm0.10$& $0.681\pm0.09$& $0.679\pm0.09$ & $0.768\pm0.04$  \\ 
SODEN \cite{tang2022soden} & $0.673\pm0.09$& $0.707\pm0.05$& $0.703\pm0.04$& $0.721\pm0.05$& $0.676\pm0.05$& $0.691\pm0.03$& $0.685\pm0.04$& $0.698\pm0.04$ & $0.710\pm0.05$ \\ 
\bottomrule
\end{tabular}
}
\end{table}

\textbf{Results on \textit{Dataset-1}: } An eye-level stratified five-fold cross-validation was performed. In each fold, the training set was further sub-divided to use $20\%$ as a validation set. %to monitor training. 
The test set in each fold comprised 667-707 scans from 47 eyes with 8 converters. While the converted dAMD scans were also used during training, they were removed from the test set to focus on forecasting conversions from iAMD images alone. The Area under the ROC curve (AUROC) and Balanced Accuracy (B.Acc) were reported for predicting %for iAMD to dAMD 
conversion within the next $6$, $12$ and $24$ months (Table 1). Concordance Index (CcI) was used to evaluate the risk scores $r$ on their ability to provide a reliable ranking of the conversion time. %%The results  are discussed below.
%\ref{Tab:dataset1} are discussed below.
% CcI is the fraction of the number of concordant image pairs (for which $r$ provides correct ranking) out of the total number of concordant pairs in the dataset that can be ranked.

\textit{Ablation Experiments} show that training with $\mathcal{L}_{cls}$ and $\mathcal{L}_{rnk}$ (row 2) leads to a marginal improvement over training with  $\mathcal{L}_{cls}$ alone (row 1) across all time-points in terms of AUROC, B.Acc(except $t=24$) as well as CcI (0.740 to 0.752) demonstrating the positive impact of imposing rank ordering. The proposed method additionally incorporates  $\mathcal{L}_{cns}$ (over row 2) which led to a considerable performance improvement in terms of CcI (0.752 to 0.783) as well as the AUROC and B.Acc across all $t$ (except for B.Acc at $t=6$).  Overall, the results demonstrate the value of using all loss terms.

\textit{Comparison with State-of-the-art} was performed against popular survival analysis techniques in rows 3-7. These include discrete survival analysis methods utilizing censored cross-entropy loss (Cens. CE) from \cite{wulczyn2020deep} and a logistic hazard model~\cite{rivail2023deep}, both employing discrete 6-month time-windows for predicting conversion. Additionally, DeepSurv \cite{katzman2018deepsurv} extends the CoxPH model using Deep Learning, while SODEN \cite{tang2022soden} is a Neural-ODE based approach, originally explored for tabular data.  These methods were implemented with the ConvNeXt-Tiny encoder by modifying the classification layers and losses.  All of these methods do not employ intra-subject regularization, hence require training a single branch network.  SODEN showed signs of overfitting with good performance on the validation set (to select the best-performing models) but led to a drop in performance on the test sets in all folds. 
The results demonstrate the superior performance of our proposed method, outperforming all other methods across all time-intervals.

\setlength{\tabcolsep}{6pt}
\renewcommand{\arraystretch}{1.2}
\begin{table}[]
\centering
\label{Tab:dataset2}
\caption{Performance (mean $\pm$ std. dev.) comparison between unsupervised (Unsup.-F) and Supervised (Sup.-F) Fine-tuning on \textit{Dataset-2}.}
\resizebox{1.0 \textwidth}{!}{
\begin{tabular}{@{}l|llHl|llHl|l@{}}
\toprule
  & \multicolumn{4}{c|}{AUROC}   & \multicolumn{4}{c|}{Balanced Accuracy} &  \\ 
                                 & \multicolumn{1}{c}{6} & \multicolumn{1}{c}{12} & \multicolumn{1}{H}{18} & \multicolumn{1}{c|}{24} & \multicolumn{1}{c}{6} & \multicolumn{1}{c}{12} & \multicolumn{1}{H}{18} & \multicolumn{1}{c|}{24} & \multicolumn{1}{c}{CcI} \\ 
\midrule
Cross-Test  & $0.748\pm0.04$& $0.764\pm0.05$& $0.767\pm0.05$& $0.758\pm0.05$& $0.702\pm0.03$& $0.712\pm0.04$& $0.715\pm0.04$& $0.707\pm0.04$ & $0.756\pm0.04$ \\ \midrule
\multicolumn{10}{c}{Finetuning with $25\%$ training data } \\  [-1ex]
\midrule
Unsup.-F  & $0.823\pm0.01$& $0.837\pm0.01$& $0.838\pm0.01$& $0.826\pm0.01$& $0.774\pm0.01$& $0.783\pm0.02$& $0.781\pm0.01$& $0.764\pm0.01$ & $0.818\pm0.01$   \\ 
Sup.-F  & $0.824\pm0.02$& $0.837\pm0.01$& $0.84\pm0.01$& $0.825\pm0.01$& $0.766\pm0.02$& $0.776\pm0.01$& $0.778\pm0.01$& $0.766\pm0.01$ & $0.816\pm0.02$   \\ 

\midrule

\multicolumn{10}{c}{Finetuning with $100\%$ training data } \\ [-1ex]
\midrule
Unsup.-F  & $0.837\pm0.01$& $0.849\pm0.01$& $0.849\pm0.01$& $0.834\pm0.01$& $0.8\pm0.01$& $0.809\pm0.01$& $0.796\pm0.01$& $0.775\pm0.01$ & $0.828\pm0.01$   \\ 
Sup.-F  & $0.845\pm0.01$& $0.853\pm0.01$& $0.856\pm0.01$& $0.845\pm0.01$& $0.786\pm0.02$& $0.793\pm0.02$& $0.792\pm0.01$& $0.773\pm0.01$ & $0.831\pm0.01$  \\ 

\bottomrule
\end{tabular}
}
\end{table}

\textbf{Results on \textit{Dataset-2}: }
We analyzed the effect of adapting our models pre-trained on  \textit{Dataset-1}, to \textit{Dataset-2} through unsupervised fine-tuning (Table 2). While the validation and test sets remained fixed across all experiments, the training set size for fine-tuning was varied by considering the entire ($100\%$) and a $25\%$ subset (Supplemental Table 4 reports $50\%$ and $75\%$). Each experiment was repeated five times, each time using a different model weight for initialization trained on each of the five-folds in  \textit{Dataset-1}. A different randomly selected subset of the training data of  \textit{Dataset-2} was employed each time except when fine-tuning on the entire ($100\%$) training dataset.
%The average performance of the 5 fine-tuned models on the held out test set is reported in Table 2. 
\textit{Cross-testing} performance in row 1, directly applied the models trained on $\textit{Dataset-1}$ without fine-tuning. A moderate drop in performance was observed in comparison to fine-tuned models, which is expected due to the image domain shift across scanners.

\begin{figure}[]
 \centering
  \includegraphics[width=1.0\textwidth]{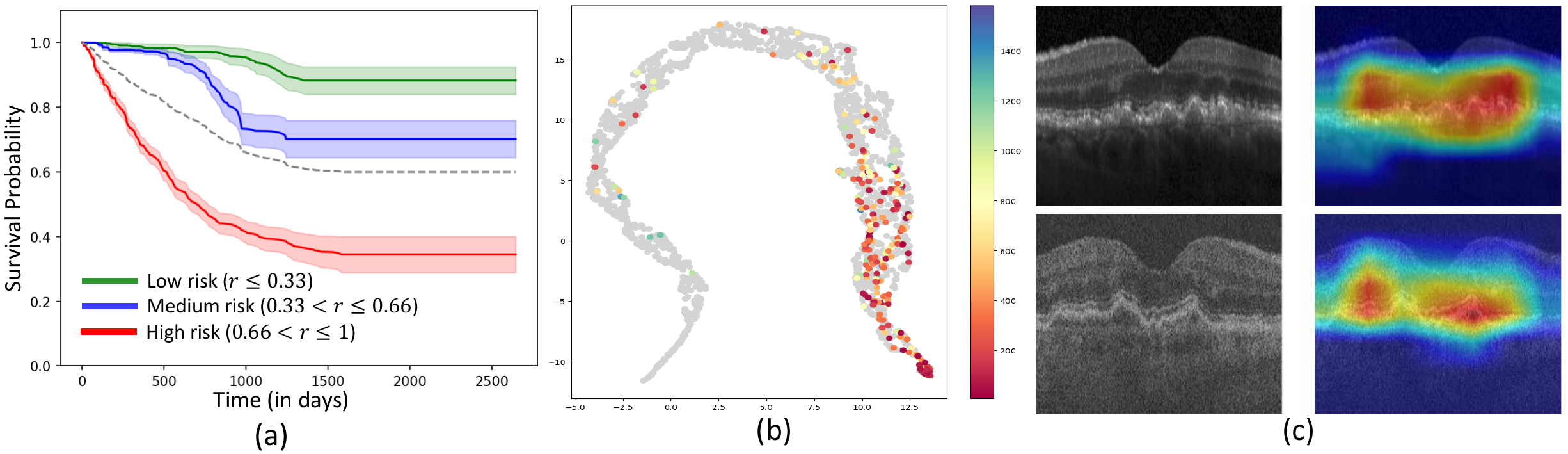}
\caption{(a) Kaplan-Meier Curves. (b) U-Map feature space visualization of the test set of $\textit{Dataset-2}$. (c) An example of domain-shift between scans from \textit{Dataset-1}(top) and \textit{Dataset-2}(bottom) along with GradCam saliency maps highlighting drusen.}
\label{fig:qual} 
 \end{figure}

\textit{Unsupervised Fine-tuning} (Unsup.-F) was performed by leveraging the inter-dependencies between longitudinal intra-subject image-pairs without using the GT training conversion-time labels. A drastic performance improvement was observed over cross-testing both in AUROC and B.Acc. across all time-intervals. The CcI improved from $0.756$ to $0.818$ by just utilizing $25\%$ of the training data (row 1 vs 2). The unsupervised fine-tuning performance further improved (row 2 vs 4) by utilizing the entire training dataset in an unsupervised manner. Fully supervised fine-tuning (Sup.-F) with GT conversion labels serve as an upper limit on fine-tuning performance. Interestingly, the performance gap between Unsup.-F and Sup-F was not significant in the small training data-regime (row 2 vs 3) with an almost same mean AUROC across $6$, $12$ and $18$ months, while Unsup.-F surpassed Sup.-F in terms of B.Acc at $t=6, 12$ and CcI (0.818 vs. 0.816). However, this trend reverses when the entire training dataset was utilized (row 4-5) in terms of AUROC, with the Unsup.-F still giving competitive performance in terms of B.Acc and CcI (0.828 for Unsup.-F compared to 0.831 for Sup.-F). Fig. \ref{fig:qual}(a) displays Kaplan-Meier survival curves for risk groups identified by thresholding $r$ from a model trained on $25\%$ of \textit{Dataset-2}'s training data with Unsup.-F. The curves %for low, medium, and high risk, based on GT labels, 
are distinctly separated, affirming $r$'s efficacy in stratifying risk. Fig. \ref{fig:qual}(b) illustrates the U-map visualization of the model's feature space, transitioning smoothly from red (short conversion time) to blue (long conversion time) along the manifold. GradCam maps in Fig. \ref{fig:qual}(c) reveal that the trained models attend to irregularities around drusen, known markers of AMD.

%%%%%%%%%%%%%%%%%%%%%%%%%%%%%%%%%%%%%%%%%%%%%%%%%%%%%%%%%%%%%%%%%%%%%%%%%%%%%%%%%%%%%

\section{Conclusion}
%Forecasting the future risk of disease progression is crucial for prioritizing high-risk patients for personalized treatment or clinical trial recruitment. %However, this task is challenging due to significant inter-subject variation in disease progression speed and a lack of established clinical biomarkers in the iAMD stage for predicting dAMD conversion. 
We proposed a novel framework to jointly predict a risk score and predict the CDF of conversion at given continuous time-intervals. The risk score, based on the signed distance of a sample from a decision hyperplane $H$ separating iAMD and dAMD samples, incorporates a ranking loss to to ensure that samples closest to $H$ have the shortest conversion time and vice versa. This temporal ordering in the feature space is further utilized to model the CDF, predicting conversion probabilities at arbitrary future time intervals using a family of hyperplanes parallel to $H$. We also enforce dependencies between intra-subject longitudinal image pairs to regularize the feature space, facilitating unsupervised fine-tuning on new datasets. Our method outperforms several popular survival analysis methods, demonstrating its effectiveness. In addition, unsupervised fine-tuning significantly improved cross-testing performance across datasets %and competitive performance with supervised fine-tuning, 
particularly with limited training data availability. This approach allows for model adaptation across datasets with significant domain shifts due to inter-scanner variability without the need for manual annotation of training labels. Future work could include evaluating our method on public datasets for related tasks like Alzheimer's disease progression from brain MRI and incorporating Longitudinal-Mixup \cite{zeghlache2023lmt} in our training to improve performance.

\begin{credits}
\subsubsection{\ackname} This research was funded in part by the Austrian Science Fund (FWF) [10.55776/FG9], and Wellcome Trust Collaborative Award Ref. 210572/Z/18/Z.

\end{credits}

\bibliographystyle{splncs04}
\bibliography{miccai_conference}

\clearpage

\appendix
\section{Supplementary Material}\label{Sec:app}

\setlength{\tabcolsep}{6pt}
\renewcommand{\arraystretch}{1.2}
\begin{table}[ht]
\centering
\label{Tab:btstrp_dataset1}
\caption{Performance at \textit{Eye-level} on \textit{Dataset-1} achieved through \textit{bootstrapping} with $ 1000$ resamplings. Each resampling involves randomly selecting one scan per eye from a single visit. Mean and standard deviation are computed across $5$ folds $\times$ $1000$ bootstrap samples.}
\resizebox{1.0 \textwidth}{!}{
\begin{tabular}{@{}l|llHl|llHl|l@{}}
\toprule
  & \multicolumn{4}{c|}{AUROC}   & \multicolumn{4}{c|}{Balanced Accuracy} &  \\ 
                                 & \multicolumn{1}{c}{6} & \multicolumn{1}{c}{12} & \multicolumn{1}{H}{18} & \multicolumn{1}{c|}{24} & \multicolumn{1}{c}{6} & \multicolumn{1}{c}{12} & \multicolumn{1}{H}{18} & \multicolumn{1}{c|}{24} & \multicolumn{1}{c}{CcI} \\ \midrule
%\multicolumn{10}{c}{Ablation} \\
%\midrule
%$\mathcal{L}_{cls}$  & $0.846\pm0.13$& $0.802\pm0.11$& $0.773\pm0.09$& $0.772\pm0.08$& $0.850\pm0.11$& $0.783\pm0.1$& $0.758\pm0.08$& $0.751\pm0.07$ & $0.752\pm0.06$  \\

%$\mathcal{L}_{cls} + \mathcal{L}_{rnk}$ & $0.854\pm0.13$& $0.808\pm0.12$& $0.778\pm0.1$& $0.774\pm0.09$& $0.858\pm0.12$& $0.807\pm0.1$& $0.768\pm0.08$& $0.765\pm0.07$ & $0.759\pm0.07$  \\ 
%\midrule
Proposed  & $\mathbf{0.835\pm0.16}$& $\mathbf{0.826\pm0.11}$& $\mathbf{0.805\pm0.08}$& $\mathbf{0.798\pm0.08}$& $\mathbf{0.824\pm0.10}$& $\mathbf{0.794\pm0.09}$& $\mathbf{0.780\pm0.08}$& $\mathbf{0.774\pm0.08}$ & $\mathbf{0.780\pm0.06}$   \\ \midrule

%\multicolumn{10}{c}{Eye-level Bootstrap performance.} \\ \midrule

%Proposed-MIDL & $0.863\pm0.1$ & $0.827\pm0.1$ & $0.808\pm0.07$ & $0.816\pm0.07$ & $0.871\pm0.11$ & $0.811\pm0.09$ & $0.789\pm0.07$ & $0.801\pm0.06$ & $0.769\pm0.06$ \\

Cens. CE \cite{wulczyn2020deep}  & $0.775\pm0.14$& $0.772\pm0.13$& $0.773\pm0.10$& $0.790\pm0.08$& $0.804\pm0.11$& $0.756\pm0.11$& $0.742\pm0.07$& $0.746\pm0.06$ & $0.762\pm0.06$   \\ 

Logis. Hazard \cite{rivail2023deep}  & $0.769\pm0.19$& $0.768\pm0.12$& $0.763\pm0.09$& $0.786\pm0.08$& $0.792\pm0.14$& $0.760\pm0.11$& $0.749\pm0.08$& $0.766\pm0.08$ & $0.749\pm0.08$ \\ 
 
DeepSurv  \cite{katzman2018deepsurv}  & $0.769\pm0.18$& $0.710\pm0.16$& $0.712\pm0.14$& $0.723\pm0.14$& $0.749\pm0.17$& $0.689\pm0.12$& $0.682\pm0.12$& $0.686\pm0.12$ & $0.752\pm0.07$ \\ 

SODEN \cite{tang2022soden}  & $0.675\pm0.24$& $0.674\pm0.17$& $0.673\pm0.13$& $0.698\pm0.11$& $0.711\pm0.19$& $0.671\pm0.14$& $0.665\pm0.11$& $0.693\pm0.10$ & $0.673\pm0.09$\\ 
\bottomrule
\end{tabular}
}
\end{table}

\setlength{\tabcolsep}{6pt}
\renewcommand{\arraystretch}{1.2}
\begin{table}[htbp]
\centering
\label{Tab:dataset2_suppl}
\caption{Performance (mean $\pm$ std. dev.) comparison between unsupervised (Unsup.-F) and Supervised (Sup.-F) Fine-tuning on \textit{Dataset-2}. }
\resizebox{1.0 \textwidth}{!}{
\begin{tabular}{@{}l|llHl|llHl|l@{}}
\toprule
  & \multicolumn{4}{c|}{AUROC}   & \multicolumn{4}{c|}{Balanced Accuracy} &  \\ 
                                 & \multicolumn{1}{c}{6} & \multicolumn{1}{c}{12} & \multicolumn{1}{H}{18} & \multicolumn{1}{c|}{24} & \multicolumn{1}{c}{6} & \multicolumn{1}{c}{12} & \multicolumn{1}{H}{18} & \multicolumn{1}{c|}{24} & \multicolumn{1}{c}{CcI} \\ 
\midrule

\multicolumn{10}{c}{Finetuning with $50\%$ training data } \\ 
\midrule
Unsup.-F & $0.829\pm0.01$& $0.843\pm0.01$& $0.845\pm0.01$& $0.829\pm0.01$& $0.790\pm0.01$& $0.801\pm0.01$& $0.794\pm0.01$& $0.772\pm0.01$ & $0.822\pm0.01$ \\
Sup.-F  & $0.835\pm0.02$& $0.847\pm0.01$& $0.851\pm0.01$& $0.839\pm0.01$& $0.788\pm0.01$& $0.795\pm0.01$& $0.790\pm0.01$& $0.772\pm0.01$ & $0.825\pm0.01$ \\ 

\midrule

\multicolumn{10}{c}{Finetuning with $75\%$ training data } \\ 
\midrule
Unsup-F  & $0.833\pm0.01$& $0.851\pm0.01$& $0.851\pm0.01$& $0.834\pm0.01$& $0.785\pm0.01$& $0.807\pm0.01$& $0.797\pm0.01$& $0.773\pm0.01$ & $0.831\pm0.01$    \\ 
Sup-F  & $0.844\pm0.01$& $0.856\pm0.01$& $0.858\pm0.01$& $0.843\pm0.01$& $0.777\pm0.01$& $0.789\pm0.01$& $0.786\pm0.01$& $0.77\pm0.01$ & $0.833\pm0.01$ \\ 

\bottomrule
\end{tabular}
}
\end{table}

\setlength{\tabcolsep}{6pt}
\renewcommand{\arraystretch}{1.2}
\begin{table}[htbp]
\centering
\label{Tab:btstrp_dataset2}
\caption{ \textit{Eye-level} Performance on \textit{Dataset-2} computed through \textit{bootstrapping} with $1000$ resamplings. Each resampling involves randomly selecting one scan per eye from a single visit. Mean and standard deviation are computed across $5$ folds $\times$ $1000$ bootstrap samples. Unsupervised(Unsup.-F) and Supervised(Sup.-F) Fine-tuning are compared.}
\resizebox{1.0 \textwidth}{!}{
\begin{tabular}{@{}l|llHl|llHl|l@{}}
\toprule
  & \multicolumn{4}{c|}{AUROC}   & \multicolumn{4}{c|}{Balanced Accuracy} &  \\ 
                                 & \multicolumn{1}{c}{6} & \multicolumn{1}{c}{12} & \multicolumn{1}{H}{18} & \multicolumn{1}{c|}{24} & \multicolumn{1}{c}{6} & \multicolumn{1}{c}{12} & \multicolumn{1}{H}{18} & \multicolumn{1}{c|}{24} & \multicolumn{1}{c}{CcI} \\ 
\midrule
Cross-Test  & $0.749\pm0.06$& $0.762\pm0.06$& $0.761\pm0.06$& $0.753\pm0.06$& $0.716\pm0.05$& $0.719\pm0.05$& $0.713\pm0.05$& $0.703\pm0.05$ & $0.739\pm0.06$   \\ \midrule
\multicolumn{10}{c}{Finetuning with $25\%$ training data } \\ 
\midrule
Unsup-F  & $0.816\pm0.03$& $0.832\pm0.02$& $0.833\pm0.02$& $0.823\pm0.02$& $0.758\pm0.03$& $0.775\pm0.03$& $0.774\pm0.02$& $0.763\pm0.02$ & $0.813\pm0.02$    \\ 
Sup-F  & $0.812\pm0.03$& $0.833\pm0.02$& $0.834\pm0.02$& $0.821\pm0.02$& $0.762\pm0.04$& $0.776\pm0.03$& $0.774\pm0.03$& $0.76\pm0.02$ & $0.81\pm0.02$   \\ 
 
\midrule

\multicolumn{10}{c}{Finetuning with $50\%$ training data } \\ 
\midrule
Unsup-F  & $0.821\pm0.03$& $0.835\pm0.02$& $0.836\pm0.02$& $0.824\pm0.02$& $0.77\pm0.03$& $0.789\pm0.02$& $0.785\pm0.02$& $0.771\pm0.02$ & $0.816\pm0.02$ \\
Sup-F  & $0.824\pm0.03$& $0.842\pm0.02$& $0.846\pm0.02$& $0.836\pm0.02$& $0.775\pm0.03$& $0.79\pm0.02$& $0.789\pm0.02$& $0.774\pm0.02$ & $0.818\pm0.02$    \\ 

\midrule

\multicolumn{10}{c}{Finetuning with $75\%$ training data } \\ 
\midrule
Unsup-F  & $0.827\pm0.03$& $0.843\pm0.02$& $0.841\pm0.02$& $0.827\pm0.02$& $0.77\pm0.03$& $0.791\pm0.02$& $0.785\pm0.02$& $0.769\pm0.02$ & $0.824\pm0.01$    \\ 
Sup-F  & $0.83\pm0.02$& $0.85\pm0.02$& $0.851\pm0.02$& $0.839\pm0.02$& $0.771\pm0.03$& $0.792\pm0.02$& $0.786\pm0.02$& $0.771\pm0.02$ & $0.827\pm0.01$ \\ 

\midrule

\multicolumn{10}{c}{Finetuning with $100\%$ training data } \\ 
\midrule
Unsup-F  & $0.838\pm0.03$& $0.847\pm0.02$& $0.846\pm0.02$& $0.834\pm0.02$& $0.788\pm0.03$& $0.796\pm0.02$& $0.787\pm0.02$& $0.772\pm0.02$ & $0.827\pm0.01$  \\ 
Sup-F  & $0.83\pm0.02$& $0.851\pm0.02$& $0.853\pm0.02$& $0.844\pm0.02$& $0.77\pm0.03$& $0.796\pm0.02$& $0.795\pm0.02$& $0.782\pm0.02$ & $0.828\pm0.01$   \\

\bottomrule
\end{tabular}
}
\end{table}

\end{document}